%% file: example_paper.tex
\documentclass{article}

\usepackage{microtype}
\usepackage{graphicx}
\usepackage{subcaption}
\usepackage{booktabs} 
\usepackage{multirow} 

\usepackage{hyperref}

\usepackage[preprint]{icml2026}

\usepackage{amsmath}
\usepackage{amssymb}
\usepackage{mathtools}
\usepackage{amsthm}

\usepackage{placeins}
\usepackage{tabularx}

\usepackage[capitalize,noabbrev]{cleveref}

\theoremstyle{plain}

\theoremstyle{definition}

\theoremstyle{remark}

\usepackage[textsize=tiny]{todonotes}

\icmltitlerunning{LQA: A Lightweight Quantized-Adaptive Framework for Vision-Language Models on the Edge}

\begin{document}

\twocolumn[
  \icmltitle{LQA: A Lightweight Quantized-Adaptive Framework for Vision-Language Models on the Edge}



  \icmlsetsymbol{equal}{*}

  \begin{icmlauthorlist}
    \icmlauthor{Xin Wang}{equal,unimelb}
    \icmlauthor{Hong Jia}{equal,uoa,unimelb}
    \icmlauthor{Hualin Zhou}{uom}
    \icmlauthor{Sheng Guang Wang}{unimelb}
    \icmlauthor{Yu Zhang}{uom}
    \icmlauthor{Ting Dang}{unimelb}
    \icmlauthor{Tao Gu}{uom}
  \end{icmlauthorlist}
  
  \icmlaffiliation{unimelb}{School of Computing and Information Systems, University of Melbourne, Melbourne, Australia}
  \icmlaffiliation{uoa}{Faculty of Science, Auckland, New Zealand}
  \icmlaffiliation{uom}{School of Computing, University of Macquarie, Sydney, Australia}

  \icmlcorrespondingauthor{Hong Jia}{hong.jia@auckland.ac.nz}

  \icmlkeywords{Machine Learning, ICML}

  \vskip 0.3in
]

\printAffiliationsAndNotice{\icmlEqualContribution}



\begin{abstract}
Deploying Vision-Language Models (VLMs) on edge devices is challenged by resource constraints and performance degradation under distribution shifts. While test-time adaptation (TTA) can counteract such shifts, existing methods are too resource-intensive for on-device deployment. To address this challenge, we propose LQA, a lightweight, quantized-adaptive framework for VLMs that combines a modality-aware quantization strategy with gradient-free test-time adaptation. We introduce Selective Hybrid Quantization (SHQ) and a quantized, gradient-free adaptation mechanism to enable robust and efficient VLM deployment on resource-constrained hardware. Experiments across both synthetic and real-world distribution shifts show that LQA improves overall adaptation performance by 4.5\%, uses less memory than full-precision models, and significantly outperforms gradient-based TTA methods, achieving up to 19.9$\times$ lower memory usage across seven open-source datasets. These results demonstrate that LQA offers a practical pathway for robust, privacy-preserving, and efficient VLM deployment on edge devices.
\end{abstract}

\input{sections/01-introduction}
\input{sections/02-related}
\input{sections/03-methods}
\input{sections/04-experiments}
\input{sections/05-results}

\section{Conclusion}
In this paper, we propose a quantizd-adaptive framework for vision-language models for edge devices. Extensive experiments show that proposed method LQA consistently improves adaptation performance, reduces memory consumption, and surpasses traditional gradient-based TTA methods in both efficiency and scalability. Our results show that LQA is a practical and effective solution for robust, privacy-preserving VLM adaptation on edge devices, paving the way for broader and more accessible real-world applications.

\bibliography{example_paper}
\bibliographystyle{icml2026}

\input{sections/Appendix}

\end{document}

%% file: sections/01-introduction.tex
\section{Introduction}
Vision--Language Models (VLMs) such as CLIP~\cite{radford2021learning} and ALIGN~\cite{jia2021scaling} deliver impressive zero--shot performance on image recognition, retrieval, and visual question answering by jointly embedding images and text. Their success has sparked interest in on-device deployment, for example on smartphones, drones, or augmented-reality headsets, where privacy, connectivity, and real-time latency favor local inference~\cite{li2024flexnn, 9711068, s23218698}.

Howevere, VLMs are highly sensitive to \emph{distribution shift}: the test images that arise in the wild often differ from those seen during pre-training, leading to pronounced accuracy drops. Distribution shift occurs whenever the statistics of the inputs observed at deployment diverge from those seen during training; for example, a user’s smartphone photo taken under dim café lighting differs markedly from the well-lit images that populate large-scale web corpora. Therefore, on-device adaptation of VLMs is essential for real-world applications. Test-Time Adaptation (TTA) was proposed to achieve adaptation by updating the model online using only the incoming unlabeled stream~\cite{wang2021tent,niu2023stabletta}. State-of-the-art (SOTA) algorithms like Tent~\cite{wang2021tent}, EATA~\cite{niu2022eata}, and the more recent RealisticTTA~\cite{su2024realistictta} have demonstrated strong performance, but they depend heavily on backpropagation via the entire network, which consumes substantial memory and makes them impractical for deployment on edge devices.

Therefore, deploying TTA for on-device VLMs is particularly challenging, as running VLMs on devices already requires significant memory and computational resources, and adding TTA for on-device adaptation further increases these resource demands. For example, 
a full-precision CLIP ViT-B/16 consumes around 350 ms and 2.7 J per image on a Snapdragon 8 Gen 3 mobile GPU; the same model on an AR headset–class Nreal X ASIC yields only 5 fps at 3 W—far from real-time~\cite{vasu2024mobileclip, qualcomm2023clip}. 
Memory is likewise tight: the 355 MB weight footprint exceeds the budget of many wearable SoCs. This consumption is solely associated with the forward pass, resulting from the large number of parameters and the use of full precision in VLMs. When applying TTA for these VLMs on-device, there are additional costs due to backpropagation. This process typically involves computing gradients across the entire network and backpropagating through all layers, which significantly increases memory usage, latency, and power consumption. Although some backpropagation-free TTA methods have been developed that do not require backpropagation
, they still rely on algorithms that optimize certain parameters which incurs computational overhead~\cite{10.5555/3692070.3693623}, or they maintain high precision for accuracy but have not addressed the practical challenges of real-world deployment~\cite{wang2021tent, su2024realistictta, niu2023sar}.

Model \emph{quantization} offers an orthogonal lever for on-device model inference: representing weights with few-bit integers shrinks the model and accelerates inference~\cite{quantsurvey, badri2023hqq}. While quantization has been extensively studied for general model inference~\cite{jacob2018integer, gholami2021survey, liu2025survey}
, its integration with TTA remains largely unexplored due to unique challenges. The effectiveness of most TTA methods relies heavily on backpropagation gradients, but low-precision arithmetic in quantized models increases both the computational cost and instability of backpropagation, often resulting in significant performance degradation~\cite{10.5555/3692070.3693623}
. For backpropagation-free TTA methods, such as the gradient-free adapters in TDA~\cite{karmanov2024TDA}, it is still an open question how reliably the adapter parameters can be optimized when using these algorithms on a quantized backbone. Currently, there is no end-to-end TTA solution that is both robust to distribution shifts and suitable for deployment on resource-constrained devices. 


This study aims to fill this gap with the proposed LQA, the first framework that \emph{jointly} addresses quantization and TTA for VLMs. LQA couples a modality-aware quantized CLIP backbone with a fully gradient-free adaptation mechanism, yielding the first VLM that is simultaneously low-precision, training-free, and robust under real-world distribution shifts for resource-constrained settings. Specifically, we proposed a selective hybrid quantization scheme: i) it achieves modality-aware compression where the vision and text encoders compress differently, with assigns an 4-bit format to the vision pathway and a 8-bit format to the text pathway; ii) we design a selective quantization mechanism where only the task-related layers are quantized while preserving a handful of cross-modal layers in full precision to balance the efficiency and accuracy. Moreover, we then integrate this low-precision backbone with a cache-based, gradient-free test-time adaptation mechanism that operates entirely in INT8/INT4.
By purposely designing the backbone and the TTA with the same low-precision representation, LQA avoids costly de-quantisation and brings state-of-the-art robustness to edge devices. Our contributions are as follows:
\begin{itemize}
\item We introduce LQA, the first vision–language framework that combines modality-aware low-bit quantization with a gradient-free TTA strategy, paving pathway for more efficient on-device adaptation.
\item We propose a data-free, modality-aware quantization scheme that selects tailored bit-widths for the vision and text encoders based on their statistical characteristics.
\item Extensive experiments on seven public datasets show that it improves robustness by an average of 4.5\,pp while reducing memory usage by up to 19.9$\times$ compared to full-precision CLIP with gradient-based TTA.
\item We deployed and evaluated LQA on consumer-grade GPUs, adapting on the fly with no training data, and the results outperform their full-precision counterparts under both synthetic and real-world distribution shifts.
\end{itemize}

%% file: sections/02-related.tex
\section{Related Work}
\label{sec:related_work}

\subsection{Test-Time Adaptation}
Test-time adaptation (TTA) targets the discrepancy between training and deployment distributions by updating the model on-the-fly using the stream of unlabeled test samples. Early work, Tent~\cite{wang2021tent}, minimizes prediction entropy to align the model with the target domain. Later, EATA~\cite{niu2022eata} removes redundant and over-confident samples to reduce memory cost, while StableTTA~\cite{niu2023sar} introduces sample-wise re-weighting for stability. Another branch leverages batch-normalization statistics for instantaneous adaptation~\cite{schneider2020covshift, nado2020ptbn}.  

Recently, backpropagation-free TTA methods have been proposed and applied to vision-language models (VLMs). Test-time Prompt Tuning (TPT)~\cite{shu2022tpt} and DiffTPT~\cite{feng2023difftpt} optimize a small set of text prompts based on a single test instance. Furthermore, the state-of-the-art Training-free Dynamic Adapter (TDA)~\cite{karmanov2024TDA} eliminates the need for gradient computation, thereby making it suitable for VLMs in TTA scenarios. However, none of these methods explicitly consider quantization during TTA, which is crucial for efficient deployment on resource-constrained devices. Additionally, they do not address the challenges posed by noisy or adversarial samples that may contaminate the adaptation buffer. In contrast, our proposed approach is both backpropagation-free and quantization-aware, enabling efficient and robust adaptation during test time.

\begin{figure*}[ht]
  \vskip 0.01in
  \begin{center}
    \includegraphics[width=\textwidth]{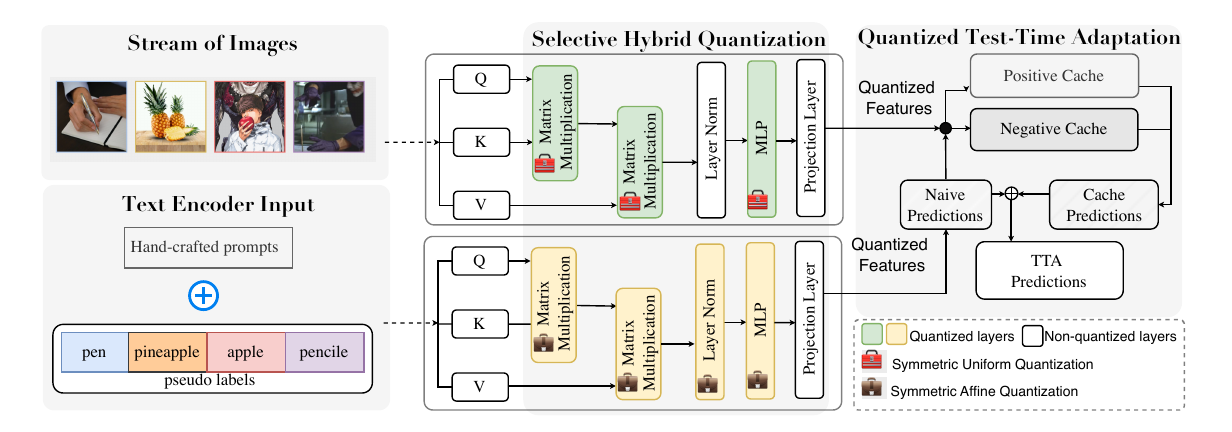}
    \caption{Diagram of the proposed LQA. (a) LQA first takes a stream of images into an image encoder along with pseudo class labels concatenated with hand crafted prompts\cite{radford2021learning}, (b) Image and text inputs are encoded by quantized CLIP backbones to generate their features, (c) Predictions are made by cosine similarity of the embeddings, (d) Image features and predictions are used to update TDA cache (e) TDA cache is applied to predictions to form final predictions.}
    \label{fig:overview}
  \end{center}
\end{figure*}

\subsection{Quantization of Vision–Language Models}
VLMs are often large and computationally intensive, making deployment on resource-limited devices challenging. 
Post-training quantization compresses neural networks by representing weights, and sometimes activations, with low-precision numbers, thereby reducing memory footprint and accelerating inference. 
Generic quantization techniques
such as bitsandbytes 8-bit quantization~\cite{dettmers2022}, GPTQ~\cite{frantar2023}, and AWQ~\cite{lin2024awq}, have proven effective for LLMs. However, these methods usually require a calibration dataset, which is often unavailable for edge deployment of VLMs. Data-free approaches, such as HQQ~\cite{badri2023hqq} and Quanto~\cite{corvoysier2024quanto}, provide more flexible fixed-point formats; however, they have not yet been explored in the context of TTA for VLMs. In contrast, our method is both adaptive and quantization-aware, while also being fully compatible with VLMs.


%% file: sections/03-methods.tex
\section{Methodology}
\label{sec:methodology}

\subsection{Problem Formulation}
Let \( \mathcal{X}_v \) and \( \mathcal{X}_t \) denote the image (visual) and text modalities, respectively. We adopt a pre-trained CLIP-style VLM comprising a vision encoder \( E_V: \mathcal{X}_v \rightarrow \mathbb{R}^d \) and a text encoder \( E_T: \mathcal{X}_t \rightarrow \mathbb{R}^d \), with feature representations compared via cosine similarity. Thses VLMs aim to classify any image into a \(\hat{y}_t \in \{1, \ldots, C\}\).

The primary objective is to adapt the model to a sequence \( \{ x_{v,t} \}_{t=1}^T \) of unlabeled images sampled from a shifted target distribution \( \mathcal{D}_T \) (\( \mathcal{D}_T \neq \mathcal{D}_{\text{train}} \)), while ensuring efficient adaptation under on-device resource constraints. Specifically, we introduce the quantization functions \( Q_V(\cdot) \) and  \( Q_T(\cdot) \) applied to the vision encoder \( E_V \) and the text encoder \( E_T \) respectively, resulting in quantized encoders \( E_V^Q = Q_V(E_V) \) and \( E_T^Q = Q_T(E_T) \) for adaptation. The aim is to optimize the quantization function \( Q_*(\cdot) \) such that the adaptation of the model to the sequence \( \{ x_{v,t} \}_{t=1}^T \) is both effective and efficient, ensuring that the total memory footprint (including model weights and cache) does not exceed a device-prescribed budget \( B_M \), and that inference for each input remains within a real-time latency requirement \( B_L \).

\subsection{System Overview}
Figure~\ref{fig:overview} shows the proposed LQA framework, which meets these constraints with a concise two-phase pipeline.
Phase~1 uses Selective Hybrid Quantization (SHQ) to compress the vision and text encoders while leaving a small set of cross-modal layers in full precision. Specifically, the quantization of $Q_V$ and $Q_T$ varies depending on the specific modality requirements, and only the layers that are most critical are optimized, while the other layers remain unchanged.

Phase~2 introduces Quantized Test-Time Adaptation (Q-TTA), a gradient-free adaptation procedure explicitly designed to operate entirely in low precision. Here, the quantization method is tailored to meet the efficiency and hardware constraints imposed by TTA, while the adaptation algorithm is, in turn, designed to be compatible with quantized representations. This joint design ensures that efficient, low-precision operation directly supports robust adaptation to distribution shifts by allowing only the vision features to be adapted, while keeping the text prototypes fixed.


\subsection{Selective Hybrid Quantization (SHQ)}
Selective Hybrid Quantization (SHQ) enables \emph{hybrid quantization} across modalities and \emph{selective precision retention} in specific fusion layers to balance accuracy and efficiency.

\subsubsection{Modality-Aware Hybrid Quantization}  Explicitly recognizes that the vision and text encoders play distinct statistical and functional roles. Specifically, the vision modality captures two-dimensional spatial information, offering a richer representation. Therefore, preserving spatial feature continuity in the vision modality is crucial. In contrast, the text modality contains one-dimensional information and offers valuable, though less detailed, data. Since the key information in text modalities resides in the token embeddings, it is especially important to leverage the distribution of these embeddings. Therefore, SHQ applies a dedicated bit-width $b_V$ or $b_T$ for vision and text modalities, where $b_V > b_T$. Furthermore, we applied a \textbf{Hessian-aware quantizer} to the vision encoder and a \textbf{symmetric uniform quantizer} to the text encoder.

SHQ independently quantizes each weight tensor \(\mathbf{W}\) in both modalities (\(\mathbf{W} \subset E_V \cup E_T\)) by determining optimal quantization parameters. Depending on the quantization type (Hessian-aware or symmetric), the optimization is slighly different. 

\noindent \textit{\underline{Hessian-aware quantization.}}
For Hessian-aware (asymmetric) quantization,  we learn both a scaling factor \(s \in \mathbb{R}_{>0}\) and an integer zero-point \(z \in \mathbb{Z}\) for each tensor.  For a given bit-width $b_V$, we seek the optimal scale $s$ and zero-point $z$ parameters that minimize the reconstruction error between the original and quantized-dequantized tensors:
\begin{equation}
(s^{\star},z^{\star}) = \underset{s>0,\,z\in\mathbb{Z}}{\arg\min}\;\|\mathbf{W} - s(\mathrm{Q}_{b}(\mathbf{W};s,z) - z)\|_2^{2},
\label{eq:scale_opt_new}
\end{equation}
where $s$ is a positive real-valued scaling factor and $z$ is the integer zero-point. The quantization operator $\mathrm{Q}_b$ maps the full-precision tensor $\mathbf{W}$ onto a low-precision $b$-bit integer range:
\begin{equation}
\mathrm{Q}_b(\mathbf{W};s,z) = \operatorname{clip}\!\bigl(\lfloor \mathbf{W}/s + z \rceil,\,0,\,2^b\!-\!1 \bigr).
\end{equation}
The output of $\mathrm{Q}_b$ is the quantized tensor, represented in integer format. To recover an approximation of the original tensor, we perform de-quantization, reconstructing the weights as:
\begin{equation}
\widetilde{\mathbf{W}} = s(\mathrm{Q}_b(\mathbf{W};s, z) - z).
\end{equation}
Here, $\mathrm{Q}_b(\mathbf{W};s,z)$ is the quantized (integer-valued) tensor, while $\widetilde{\mathbf{W}}$ is the de-quantized (floating-point) tensor, used to approximate the original full-precision weights.

In practice, the optimisation in Eq.~\eqref{eq:scale_opt_new} is approximated via a closed-form update derived from the Hessian of the per-layer reconstruction loss~\cite{frantar2023gptq}. We keep the feature dimensions unchanged between quantized and full-precision representations, i.e., $d_q = d$, but the quantized tensors are stored using $b$-bit integers instead of standard floating-point formats.

\noindent \textit{\underline{Symmetric Uniform Quantization.}}
For symmetric quantization, the zero-point \(z\) is fixed to zero, centering the quantization range around zero. The quantization step becomes:
\begin{equation}
s^* = \underset{s>0}{\arg\min}\;\|\mathbf{W} - s\,\mathrm{Q}_b(\mathbf{W}; s)\|_2^2 ,
\label{eq:scale_opt_symmetric}
\end{equation}
with the quantizer defined as:
\begin{equation}
\mathrm{Q}_b(\mathbf{W}; s) = \operatorname{clip}\left( \lfloor \mathbf{W}/s \rceil,\, -q_{\max},\, q_{\max} \right ) ,
\end{equation}
where \(q_{\max} = 2^{b-1}-1\). The dequantized tensor is then:
\begin{equation}
\widetilde{\mathbf{W}} = s \cdot \mathrm{Q}_b(\mathbf{W}; s).
\end{equation}
The remaining process for optimization is similar as Hessian-aware quantization.


\subsubsection{Selective Precision Retention}
It is important to avoid quantization-induced distortions in certain key layers specific to each modality and textual branch. To achieve this, we designate a small subset of projection layers, indexed by $\mathcal{L}$, to remain in full precision. The set $\mathcal{L}$ is selected based on a Hessian trace heuristic~\cite{dong2020hawqv2}
, which identifies layers that are most sensitive to quantization. In practice, $\mathcal{L}$ typically contains only a few (e.g., 1-2 per modality) layers for each modality, striking a balance between model compactness and the preservation of critical information flow.

Formally, for each layer $\ell$, the quantization is applied as follows:
\begin{equation}
\widetilde{\mathbf{W}}^{(\ell)} = 
\begin{cases}
    \text{dequantize}(\mathbf{W}^{(\ell)}), & \ell \notin \mathcal{S}, \\
    \mathbf{W}^{(\ell)}, & \ell \in \mathcal{S}.
\end{cases}
\end{equation}
where layers in $\mathcal{L}$ retain their original full-precision representations, while all other weights are stored in $b_V$- or $b_T$-bit quantized format. This selective protection of key layers ensures efficient model compression without sacrificing performance.

\begin{algorithm}[t]
\caption{On-the-Fly Adaptation with LQA}
\label{alg:lqa_vlm}
\begin{algorithmic}[1]
\STATE \textbf{Input:} vision encoder $E_V$, text encoder $E_T$, bit-widths $(b_V,b_T)$, test stream $D_{\text{test}}$, cache size $K$, gains $(\lambda,\mu)$, confidence threshold $\tau$
\STATE \COMMENT{Phase~1: quantise encoders and prepare text prototypes}
\STATE $E_V^{Q}\leftarrow \text{SHQ}(E_V,b_V)$; \quad $E_T^{Q}\leftarrow \text{SHQ}(E_T,b_T)$
\STATE $P^{Q}\leftarrow E_T^{Q}(\text{prompts})$ \COMMENT{freeze quantised text prototypes}
\STATE Initialise positive cache $\mathcal{C}^{+}\leftarrow\emptyset$, negative cache $\mathcal{C}^{-}\leftarrow\emptyset$
\STATE \COMMENT{Phase~2: streaming inference + adaptation}
\FOR{each image $x$ in $D_{\text{test}}$}
    \STATE $\mathbf{f}\leftarrow E_V^{Q}(x)$ \COMMENT{INT8/INT4 vision feature}
    \STATE $\boldsymbol{\ell}\leftarrow \mathbf{f} P^{Q\top}$ \COMMENT{baseline logits}
    \STATE $\boldsymbol{\ell}\leftarrow \boldsymbol{\ell}+ \lambda\sum_{\mathbf{g}\in\mathcal{C}^{+}}\mathbf{f}^{\top}\mathbf{g}-\mu\sum_{\mathbf{g}\in\mathcal{C}^{-}}\mathbf{f}^{\top}\mathbf{g}$
    \STATE $\hat{y}\leftarrow \arg\max_c \ell_c$; \quad $p_{\text{conf}}\leftarrow \max\text{softmax}(\boldsymbol{\ell})$
    \IF{$p_{\text{conf}}\ge \tau$}
        \IF{pseudo-label of $x$ matches $\hat{y}$}
            \STATE Push $\mathbf{f}$ to $\mathcal{C}^{+}$ (truncate to $K$)
        \ELSE
            \STATE Push $\mathbf{f}$ to $\mathcal{C}^{-}$ (truncate to $K$)
        \ENDIF
    \ENDIF
    \STATE \textbf{output} $\hat{y}$
\ENDFOR
\end{algorithmic}
\end{algorithm}

\subsection{Quantized Test-Time Adaptation (Q-TTA)}

As illustrated in Figure~\ref{fig:overview}, for each input image, we first extract quantized feature vectors from the vision ($\mathbf{q}_V$) and text ($\mathbf{q}_T$) encoders. The cosine similarity between these vectors is used to produce the initial prediction: a $C$-dimensional softmax probability vector $\mathbf{p}_v$, representing the likelihood of the image belonging to each class.

To adapt and refine these predictions online, we maintain two FIFO (First-In, First-Out) memory queues of quantized exemplar features: a positive cache $\mathcal{C}^{+}$ and a negative cache $\mathcal{C}^{-}$, each containing up to $K$ exemplar features along with their associated probability vectors, $\{\mathbf{e}_{+/-}, \mathbf{p}_{+/-}\}$. For an incoming image, its quantized feature $\mathbf{q}_V$ is compared against the entries in both caches, and the computed similarities are used as adaptive weights to adjust the initial prediction.

Following TDA conventions, adaptation is performed as:
\begin{align}
\ell_c(\mathbf{q}_V) & = \mathbf{p}_v + \lambda \sum_{\mathbf{e}^+ \in \mathcal{C}^{+}} (\mathbf{q}_V^\top \mathbf{e}^+) \cdot p_{e^+, c} \\ \nonumber
& - \mu \sum_{\mathbf{e}^- \in \mathcal{C}^{-}} (\mathbf{q}_V^\top \mathbf{e}^-) \cdot p_{e^-, c}
\end{align}
where 
$\lambda$ and $\mu$ are hyperparameters controlling the influence of positive and negative caches, and $p_{e^+,c}$ ($p_{e^-,c}$) is the class-$c$ probability of the cached positive (negative) exemplar.

Throughout deployment, the cached exemplars are dynamically updated: images with low-entropy (high-confidence) predictions are treated as positives and stored in $\mathcal{C}^+$, while those with high-entropy predictions (within a specified range) are assigned to $\mathcal{C}^-$. For more information, refer to Algorithm 1. Additional details regarding cache management and entropy thresholds are provided in the Appendix.


%% file: sections/04-experiments.tex
\section{Experiments}

\label{sec:guidelines}
\subsection{Datasets} 
We evaluated our approach using a variety of benchmarking datasets. CIFAR10C and CIFAR100C \cite{cifar} are used, which contain 10 and 100 object classes respectively, each featuring 15 different artificial corruption variants applied to the test images. To assess performance under real-world distribution shift, we utilized Caltech101~\cite{caltech101}, Oxford Pets\cite{oxford_pets}, Describable Textures Dataset (DTD)\cite{dtd}, UFC101~\cite{ucf101}, and ImageNet Adversarial (ImageNet A)\cite{hendrycks2021imgA}. Details of datasets can be found in Appendix.


\subsection{Baselines}
We evaluate our proposed methods from two perspectives. First, we compare LQA against a variety of SOTA TTA baselines. Second, we evaluate the impact of various quantization strategies on accuracy, latency, and memory usage by substituting our LQA method with alternative quantization techniques for a fair comparison.
\subsubsection{TTA baselines} We evaluated our proposed LQA to other TTA methods under various settings: (i) comparison with the full-precision CLIP model without adaptation, which serves as the baseline for source model performance; (ii) comparison with SOTA full-precision TTA methods, including four approaches: Tent~\cite{wang2021tent}, which minimizes prediction entropy on the current mini-batch and updates only batch normalization parameters; EATA~\cite{niu2022eata}, which improves upon Tent by discarding low-confidence and redundant samples prior to adaptation; SAR~\cite{niu2023sar}, which augments entropy minimization with sharpness-aware regularisation and a reliability gate to prevent catastrophic drift; and TDA~\cite{karmanov2024TDA}, which employs a lightweight dynamic adapter trained on-the-fly using a nearest-neighbor cache, thereby completely avoiding back-propagation.  

\subsubsection{Quantization baselines} 
We also compared our approach to different SOTA quantization baselines.
\begin{itemize}
    \item Half-Quadratic Quantization (HQQ) \cite{badri2023hqq}: A data-free method that optimizes for quantization error using a sparsity-promoting loss. We evaluated HQQ across bit sizes (1, 2, 3, 4, 8) and group sizes (8 to 512).
    \item BitsandBytes (BnB) \cite{dettmers2022llm}: A technique combining vector-wise quantization with 16-bit matrix multiplication for outliers, tested at 4-bit and 8-bit precision.
    \item Quanto \cite{corvoysier2024quanto}: A linear quantization library that maps weights to a uniform fixed-point representation, evaluated at 2-bit and 8-bit precision. (4-bit was precluded by backend compatibility issues).
\end{itemize}

\subsection{Experiment Setting and Evaluations}
\label{sec:definition_experimemt}
All methods and experiments were evaluated using a batch size of 1 to simulate real-world data stream scenarios commonly encountered in edge device deployments. Performance was measured by classification accuracy. Additionally, efficiency was assessed using two metrics: latency, defined as the time required to process a single test sample, and memory usage, defined as the maximum GPU memory allocated per sample. Both efficiency metrics were averaged over 10 samples. All experiments were conducted on a consumer-grade NVIDIA RTX 4070 Ti SUPER GPU with 16 GB of memory. Details of the hyperparameters and software versions are provided in the Appendix.

%% file: sections/05-results.tex
\section{Results}
\label{sec:results}


\begin{table}[t]
\centering
\caption{Performance comparison of TTA on the CIFAR-10 dataset. \textbf{Best} in bold and \underline{second best} underlined.}
\label{tab:experiment_results_tta_cifar10}

\resizebox{\columnwidth}{!}{
\begin{tabular}{lccccccc}
\toprule
{} & CLIP & TENT & EATA & SAR & TDA & LQA(Ours) & LQA-Lite(Ours)  \\
\midrule
CIFAR10 & 89.25 & \textbf{92.89} & \underline{91.45} & 90.59 & 91.73 & 91.66 & 90.94 \\
\midrule
Gaussian Noise & 37.72 & 16.36 & 43.75 & \textbf{47.00} & \underline{43.96} & 43.82 & 40.33 \\
Shot Noise & 41.11 & 28.24 & 48.35 & \underline{50.09} & 48.14 & \textbf{48.59} & 43.51 \\
Impulse Noise & 51.75 & 49.93 & 59.97 & 56.66 & \textbf{60.46} & \underline{60.17} & 59.38 \\
Defocus Blur & 70.04 & \textbf{76.92} & 73.65 & \underline{75.13} & 73.84 & 73.84 & 72.69 \\
Glass Blur & 42.22 & 36.73 & 46.15 & 46.53 & \textbf{47.34} & \underline{46.38} & 46.06 \\
Motion Blur & 65.85 & \textbf{77.58} & 71.25 & 74.16 & 71.03 & \underline{71.52} & 70.06 \\
Zoom Blur & 72.60 & 76.45 & \underline{76.05} & 76.57 & 75.81 & \textbf{76.27} & 74.21 \\
Snow & 73.30 & 78.85 & \underline{77.35} & \textbf{79.22} & 77.66 & 77.53 & 75.95 \\
Frost & 76.65 & 79.20 & \underline{80.15} & 80.77 & 80.33 & \textbf{80.34} & 78.47 \\
Fog & 68.43 & 72.55 & 72.95 & \textbf{75.96} & \underline{73.27} & 73.11 & 71.60 \\
Brightness & 83.43 & 86.25 & 86.25 & \textbf{86.92} & \underline{86.68} & 86.50 & 84.82 \\
Contrast & 61.98 & 65.25 & 66.35 & \textbf{71.99} & 66.03 & \underline{66.57} & 63.81 \\
Elastic Transform & 53.15 & 55.80 & \underline{57.15} & 57.21 & 56.95 & \textbf{57.34} & 56.86 \\
Pixelate & 48.51 & \textbf{67.53} & 53.65 & \underline{64.20} & 54.79 & 53.82 & 50.41 \\
JPEG Compression & 56.09 & \textbf{66.51} & \underline{63.15} & 61.73 & 62.90 & 63.34 & 61.41 \\
\midrule
Average & 60.19 & 64.96 & 66.75 & \underline{66.94} & 65.25 & \textbf{66.95} & 65.03 \\
\bottomrule
\end{tabular}
}
\end{table}

\begin{table}[t]
\caption{Performance comparison of TTA on the CIFAR-100 dataset. \textbf{Best} in bold and \underline{second best} underlined.}
\label{tab:experiment_results_tta_cifar100}

\resizebox{\columnwidth}{!}{
\begin{tabular}{lccccccc}
\toprule
{} & CLIP & TENT & EATA & SAR & TDA & LQA(Ours) & LQA-Lite(Ours)  \\
\midrule
CIFAR100 &  64.72 &  \underline{71.61} &  \textbf{72.08} &  70.97 &  69.88 & 69.66 & 67.79 \\
\midrule
Gaussian Noise      &  15.86 &  3.52 &  21.45 &  18.67 &  \underline{24.01} & \textbf{24.13} & 20.01 \\
Shot Noise         &  17.52 &  3.39 &  24.51 &  20.53 &  \textbf{25.99} & 26.02 & 21.87 \\
Impulse Noise      &  21.33 &  5.17 &  28.43 &  23.99 &  \underline{32.37} & \textbf{32.89} & 29.19 \\
Defocus Blur      &  40.09 & \textbf{50.11} &  \underline{48.71} &  47.04 &  45.34 & 45.28 & 44.53 \\
Glass Blur        &  13.49 &  2.61 &  15.76 &  14.36 &  \underline{21.61} & 21.57 & \textbf{22.19} \\
Motion Blur       &  39.82 & \textbf{49.06} &  \underline{48.59} &  46.86 &  44.52 & 44.58 & 42.18 \\
Zoom Blur         &  45.43 & \textbf{53.99} &  \underline{53.65} &  51.58 &  50.15 & \textbf{50.50} & 48.93 \\
Snow              &  42.79 & \underline{52.94} &  \textbf{53.86} &  51.65 &  49.45 & 49.68 & 47.21 \\
Frost             &  45.41 & 45.91 &  \textbf{52.00} &  50.66 &  51.40 & \underline{51.58} & 47.98 \\
Fog               &  38.88 & \textbf{51.26} &  \underline{50.38} &  49.41 &  42.94 & 43.03 & 41.51 \\
Brightness        &  52.51 & \underline{63.77} &  \textbf{63.95} &  62.40 &  59.24 & 59.03 & 56.77 \\
Contrast          &  33.41 & \textbf{50.92} &  \underline{47.89} &  44.41 &  36.78 & 37.08 & 35.88 \\
Elastic Transform &  24.34 & \textbf{32.66} &  \underline{31.53} &  30.27 &  30.42 & 29.65 & \textbf{30.63} \\
Pixelate          &  21.91 & 24.18 &  \textbf{31.15} &  \underline{30.26} &  28.34 & 28.00 & 25.91 \\
JPEG Compression   &  27.18 & 32.33 &  \textbf{36.52} &  34.37 &  \underline{34.92} & 34.31 & 33.17 \\
\midrule
Average               &  32.00 & 34.79 &  \textbf{40.56} &  38.43 &  38.50 & 40.44 & \underline{38.48} \\
\bottomrule
\end{tabular}
}
\end{table}

\vspace{-1pt}

\subsection{Comparison with TTA Methods}
\label{sec:overall_results}
We evaluate the proposed method, LQA, against several SOTA TTA baselines. Additionally, we include a variant of LQA, called LQA-Lite, which is more efficient and uses less memory. This variant is included to assess the best trade-off between system overhead and performance.
Table~\ref{tab:experiment_results_tta_cifar10} and Table~\ref{tab:experiment_results_tta_cifar100} show the results on CIFAR-10 and CIFAR-100, respectively. On CIFAR-10, LQA achieves the best average accuracy of 66.95\%, equaling the performance of the strongest baseline, SAR (66.94\%), and exceeding that of EATA (66.75\%) and TDA (65.25\%). This pattern persists on the more challenging CIFAR-100 benchmark, where LQA obtains an accuracy of 40.44\%, closely matching the leading baseline, EATA (40.56\%). Notably, our approach exhibits pronounced robustness to noise-based corruptions. For example, on CIFAR-100 with Gaussian Noise, LQA attains an accuracy of 24.13\%, whereas TENT’s performance drops markedly to 3.52\%. The lightweight LQA-Lite variant also demonstrates strong performance, achieving accuracy comparable to TDA on CIFAR-10 (65.03\% vs.\ 65.25\%) and performing on par with both SAR (38.43\%) and TDA (38.50\%) on CIFAR-100 with an accuracy of 38.48\%.

The overall results from these two databases, as well as performance on five other real-world databases, are summarized in Table~\ref{tab:experiment_results_tta_all}. These results further reinforce the efficacy of our approach. LQA attains the highest mean accuracy (65.56\%) across all seven datasets, surpassing the second best method, TDA (64.97\%). This is underpinned by strong performance across multiple datasets, including SOTA results on Oxford Pets (89.75\%) and UCF101 (71.11\%). Although some baselines achieve competitive results on individual benchmarks, they do not demonstrate the same level of consistency as LQA. Furthermore, LQA-Lite maintains robust generalization, with an overall average of 63.98\%, exceeding baselines such as TENT (62.80\%) and SAR (63.28\%). These findings support the assertion that our quality-aware adaptation strategy provides a robust and reliable solution for real-world TTA applications.

\begin{table}[t!]
\centering
\caption{Overall performance on seven different datasets. \textbf{Best} in bold and \underline{second best} underlined.}
\label{tab:experiment_results_tta_all}

\resizebox{\columnwidth}{!}{
\begin{tabular}{lccccccc}
\toprule
{} &   CLIP &   TENT &   EATA &    SAR &    TDA &  LQA(Ours)  &  LQA-Lite(Ours)  \\
\midrule
CIFAR10C  &  60.19 &  64.96 &  66.75 &  \underline{66.94} &  65.25 & \textbf{66.95} & 65.03 \\
CIFAR100C &  32.00 &  34.79 &  \textbf{40.56} &  38.43 &  38.50 & \underline{40.44} & 38.48 \\
Caltech 101    &  91.56 &  92.17 &  92.62 &  92.37 &  \textbf{94.28} & 93.79 & \underline{93.96} \\
Dtd           &  44.56 &  45.21 &  44.74 &  44.62 &  \textbf{45.45} & \underline{45.39} & 45.15 \\
Oxford pets   &  87.16 &  87.60 &  88.06 &  87.54 &  \underline{89.62} & \textbf{89.75} & 88.25 \\
ucf101         &  63.89 &  65.00 &  65.11 &  63.94 &  \underline{70.05} & \textbf{71.11} & 68.73 \\
ImageNet A   &  47.97 &  49.84 &  49.88 &  49.10 &  \textbf{51.63} & \underline{51.46} & 48.25 \\
\midrule
Average   &  61.05 &  62.80 &  64.33 &  63.28 &  \underline{64.97} & \textbf{65.56} & 63.98 \\
\bottomrule
\end{tabular}
}

\end{table}
\vspace{-3pt}



    \begin{table*}[t!]
    \centering
    \caption{Comparison of various quantization methods across all datasets.}
    \label{tab:experiment_results_quantisation_all}
    \resizebox{2\columnwidth}{!} {
    \begin{tabular}{lcc|cccc|cccc}
    \toprule
    & CLIP & TDA & BnB(4b) & HQQ(4b) & Quanto(4b) & \textbf{LQA-Lite(Ours)} & BnB(8b) & HQQ(8b) & Quanto(8b) & \textbf{LQA(Ours)} \\
    \midrule
    CIFAR10C     & 61.99 & 66.78 & 59.28 & 64.98 & 62.70 & 65.03 & \textbf{66.99} & 66.58 & 66.91 & \underline{66.95} \\
    CIFAR100C    & 34.03 & 40.44 & 33.53 & 38.06 & 32.13 & 38.48 & \underline{40.57} & 40.35 & \textbf{40.68} & 40.44 \\
    Catechd101   & 91.48 & \textbf{94.32} & 92.21 & 92.78 & 92.33 & 93.96 & 94.16 & 94.04 & \underline{94.24} & 93.79 \\
    Dtd          & 44.62 & \textbf{46.45} & 42.55 & 43.56 & 43.68 & 45.15 & 45.80 & 45.74 & \underline{45.86} & 45.39 \\
    Oxford Pets  & 87.24 & \textbf{89.86} & 85.04 & 86.15 & 85.42 & 88.25 & 89.75 & \underline{89.83} & 89.56 & 89.75 \\
    UCF101       & 63.86 & 70.18 & 65.87 & 67.94 & 65.77 & 68.73 & 70.05 & \underline{71.00} & 70.87 & \textbf{71.11} \\
    ImageNet A   & 47.86 & 51.30 & 41.46 & 46.90 & 40.92 & 48.25 & 51.17 & 51.38 & \textbf{51.73} & \underline{51.46} \\
    \midrule
    Average      & 61.58 & 65.62 & 59.99 & 62.91 & 60.42 & 63.98 & 65.50 & 65.56 & 65.69 & 65.56 \\
    \bottomrule
    \end{tabular}
    }
    \end{table*}

\begin{figure*}[t]
    \centering
    \includegraphics[width=\linewidth]{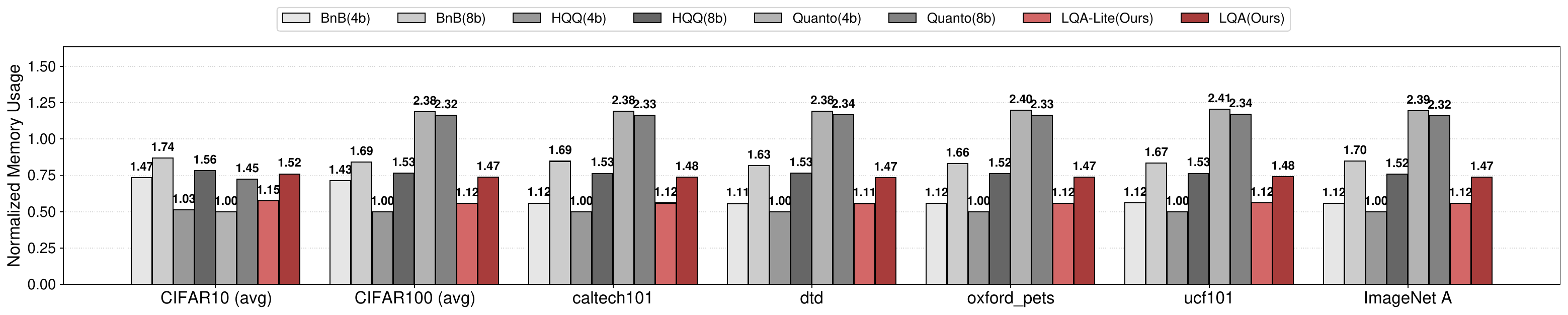}
    \caption{Comparison of runtime memory usage for different quantization methods.}
    \label{fig:memory_compare_quantization}
\end{figure*}

\begin{figure}[t!]
    \centering
    \begin{subfigure}{0.49\columnwidth}
        \includegraphics[width=\textwidth]{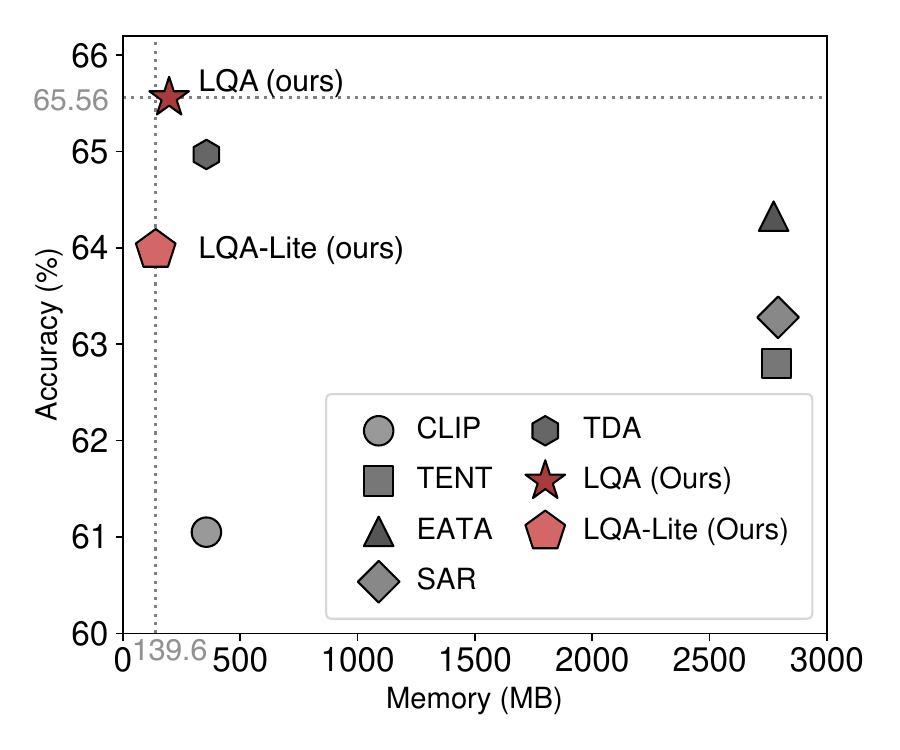}
        \caption{Memory vs Accuracy}
        \label{fig:memory_pareto}
    \end{subfigure}
    \begin{subfigure}{0.49\columnwidth}
        \includegraphics[width=\textwidth]{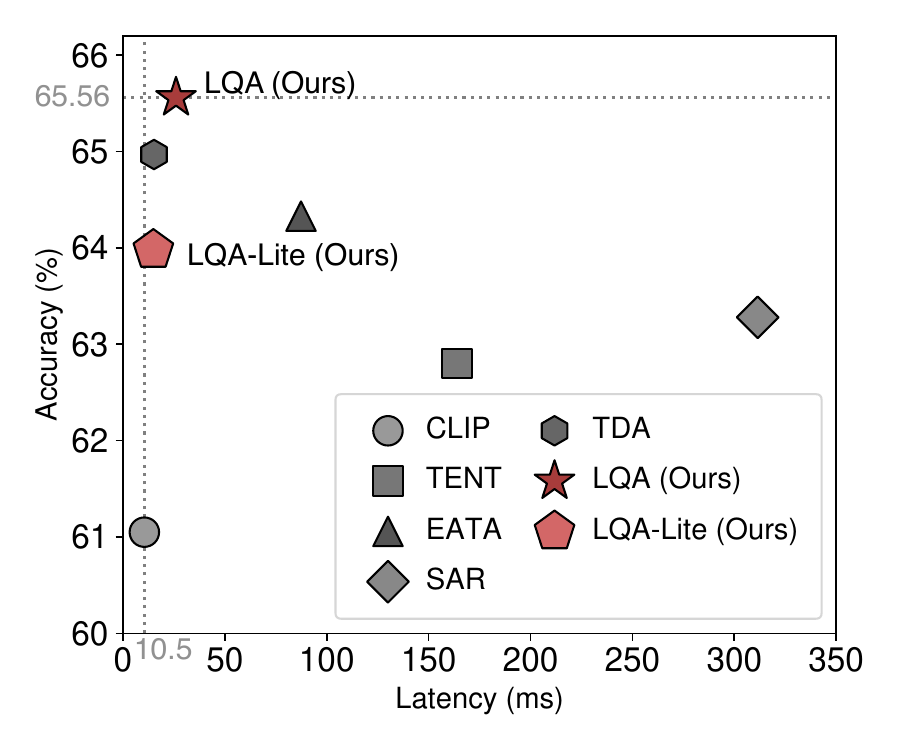}
        \caption{Latency vs Accuracy}
        \label{fig:latency_pareto}
    \end{subfigure}
    \caption{(a) Memory consumption and (b) latencies versus accuracy.}
    \raggedright
    \vspace{-2.5pt}
\end{figure}

\begin{figure*}[t]
    \centering
    \includegraphics[width=\linewidth]{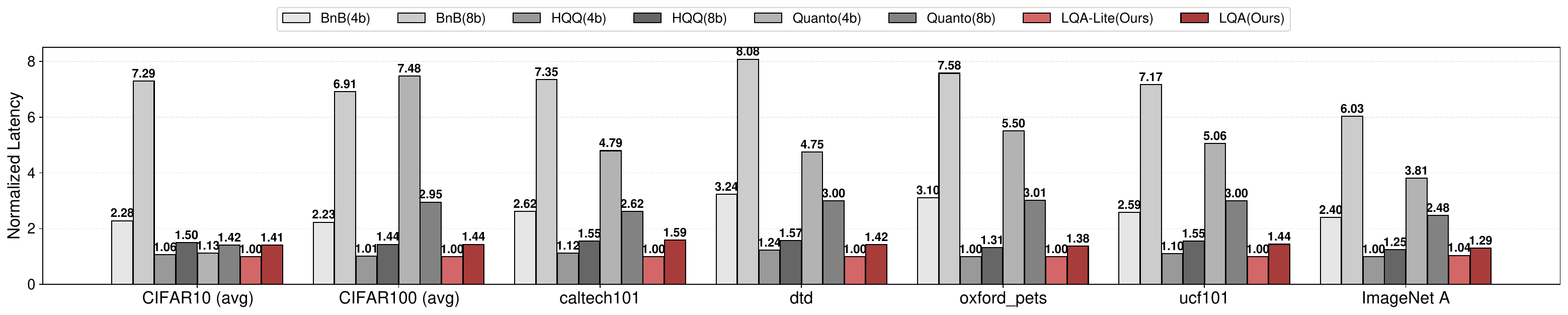}
    \caption{Comparison of runtime latency across various quantization methods.}
    \label{fig:latency_compare_quantization}
    \vspace{-1em}
\end{figure*}  


\subsection{Memory and Latency for TTA}
\label{sec:efficiency_results}
The memory usage comparison is shown in Figure~\ref{fig:memory_pareto}. Our method, LQA, achieves the highest accuracy 
with a significant reduction in memory, requiring only $196.73$ MB, which is approximately $1.8\times$ less than the $355.71$ MB used by the second-best model of TDA. Compared to other methods like EATA, which demand over $2700$ MB of memory, LQA not only offers a $1.23\%$ improvement in accuracy but also achieves a remarkable $14.1\times$ reduction in memory footprint. Furthermore, our LQA-Lite variant provides an even more memory-efficient solution. It attains a competitive accuracy 
to that of SAR and EATA, while utilizing a mere $139.63$ MB. This represents a substantial $19.9\times$ memory saving compared to EATA, highlighting the exceptional efficiency and practicality of our approach.

The evaluation of accuracy versus inference latency further demonstrates the efficiency of our proposed methods. As shown in the Figure~\ref{fig:latency_pareto}, LQA maintains  a low latency of $26.04$ ms with the best performance. 
In comparison to other methods like EATA, LQA reduces inference latency by approximately $3.4\times$ without sacrificing accuracy. Furthermore, our LQA-Lite variant is optimized for speed, achieving a latency of only $14.92$ ms, which is faster than TDA. At this speed, it delivers a competitive accuracy of $63.98\%$, marking a nearly $5.9\times$ reduction in latency compared to EATA for a minimal $0.35\%$ decrease in accuracy. This highlights the compelling trade-off our methods offer between computational speed and predictive performance.

\subsection{Comparison to SOTA Quantization Methods}
The comprehensive results across multiple datasets are presented in Table~\ref{tab:experiment_results_quantisation_all}. The analysis indicates that our proposed methods maintain strong performance across a variety of benchmarks. For 4-bit quantization, LQA-Lite achieves an average accuracy of 63.98\%, outperforming other dedicated 4-bit quantization methods. Notably, it shows a 1.07\% improvement in average accuracy over the second best method, HQQ(4b). In the 8-bit quantization setting, our LQA method demonstrates competitive performance with an average accuracy of 65.56\%. While its average score is comparable to other 8-bit methods, LQA achieves the highest accuracy on the UCF101 dataset at 71.11\% and secures the second-best performance on ImageNet A. These findings underscore the effectiveness of our approach in preserving model accuracy under quantisation for a diverse range of visual recognition tasks. Further details are provided in the Appendix.


\subsection{Memory and Latency with Quantization}
The memory usage result is shown in Figure~\ref{fig:memory_compare_quantization}, which presents the normalized memory consumption for various quantization methods. Our proposed 8-bit method, LQA, consistently demonstrates an efficient memory profile. In comparison to BnB(8b), LQA achieves an average memory reduction of approximately $12\%$ across all tested datasets. The memory savings are more pronounced when compared with Quanto(8b), where LQA reduces consumption by an average of $36.5\%$ on datasets other than CIFAR10. Our 4-bit variant, LQA-Lite, also exhibits a competitive memory footprint. It reduces memory usage by over $20\%$ compared to BnB(4b) on the CIFAR10 dataset. Furthermore, LQA-Lite achieves a substantial average memory saving of approximately $53\%$ against Quanto(4b) across all datasets except for CIFAR10, underscoring the efficiency of our approach.

The latency result is shown in Figure~\ref{fig:latency_compare_quantization}, which details the normalized speed for each quantization method. Our 4-bit approach, LQA-Lite, consistently records the lowest or nearly the lowest latency across all evaluated datasets. It offers substantial speed improvements over other methods, achieving an average speedup of more than $2.5\times$ relative to BnB(4b) and over $4.5\times$ relative to Quanto(4b). In the 8-bit category, our LQA method also demonstrates a considerable performance advantage. On average, LQA is approximately $5\times$ faster than BnB(8b) and provides a speedup of over $1.8\times$ compared to Quanto(8b). In relation to HQQ, both LQA-Lite and LQA exhibit highly competitive latency, often showing a slight performance edge. These findings indicate that our proposed methods yield significant reductions in inference time compared to several existing techniques. Detailed information regarding memory usage and latency is provided in the Appendix.

\begin{table}[t]
    \centering
    \caption{Ablation study of proposed methods.}
    \label{tab:lqa_combined_resized}
    \resizebox{\columnwidth}{!}{
        \begin{tabular}{llccc}
        \toprule
        \textbf{Method} & \textbf{Configuration} & \textbf{Accuracy (\%)} & \textbf{Latency (ms)} & \textbf{Memory (MB)} \\
        \midrule
        \multirow{3}{*}{LQA-Lite (4-bit)} & Vision Encoder Only & 64.13 & 8.9 & 184.64 \\
                                          & Text Encoder Only & 65.69 & 7.1 & 265.80 \\
                                          & Both Encoders & 63.98 & 8.1 & 148.31 \\
        \cmidrule(lr){1-5}
        \multirow{3}{*}{LQA (8-bit)}     & Vision Encoder Only & 65.89 & 11.3 & 231.08 \\
                                         & Text Encoder Only & 65.76 & 6.8 & 296.38 \\
                                         & Both Encoders & 65.56 & 11.4 & 195.76 \\
        \bottomrule
        \end{tabular}
    }
\vspace{-1.5em}
\end{table}
\vspace{-3pt}

\subsection{Ablation Study}
The ablation study results are shown in Table~\ref{tab:lqa_combined_resized}, detailing the performance of our models under different configurations and quantization levels. The analysis indicates that utilizing both the vision and text encoder generally yields a competitive balance between accuracy and computational cost, achieving the lowest latency for both the 4-bit and 8-bit models (7.1 ms and 6.8 ms, respectively). 

When comparing the quantization methods, the LQA-Lite (4-bit) model demonstrates considerable advantages in efficiency. For instance, the LQA-Lite configuration using both encoders reduces memory consumption by 47.45 MB, a 24.2\% decrease, and lowers latency by 3.3 ms relative to its 8-bit counterpart. This significant reduction in resource requirements is achieved with a modest decrease in accuracy of only 1.58\%, presenting a favorable trade-off for deployment in environments with limited computational resources. 

In addition to trade-offs analysis, we also conduct an extensive ablation study to evaluate the performance of various quantization methods at different precision levels (4-bit, 8-bit, and low-bits) across three encoder configurations: vision encoder only, text encoder only, and both encoders. The results are summarized in Tables~\ref{tab:ablation-main} and~\ref{tab:ablation-lowbit}, reporting accuracy, runtime latency, and memory usage for each setting.

\begin{table}[t]
\centering
\caption{Ablation study of various quantization methods at 4-bit and 8-bit precision, evaluated on different encoder configurations.}
\resizebox{\columnwidth}{!}{
\begin{tabular}{lccccc}
\toprule
\textbf{Method} & \textbf{Vision} & \textbf{Text} & \textbf{Accuracy (\%)} & \textbf{Latency (sec)} & \textbf{Memory (MB)} \\
\midrule
BnB (4b)         & \checkmark & --         & 61.45 & 0.0216 & 198.69 \\
HQQ (4b)         & \checkmark & --         & 64.13 & 0.0092 & 184.75 \\
Quanto (4b)         & \checkmark & --         & 61.12 & 0.0345 & 298.40 \\
LQA-Lite(Ours)         & \checkmark & --         & 64.13 & 0.0089 & 184.64 \\

BnB (8b)         & \checkmark & --         & 65.51 & 0.0561 & 252.64 \\
HQQ (8b)         & \checkmark & --         & 65.31 & 0.0119 & 232.19 \\
Quanto (8b)         & \checkmark & --         & 65.80 & 0.0215 & 296.65 \\
LQA(Ours)         & \checkmark & --         & 65.89 & 0.0113 & 231.08 \\
\midrule
BnB (4b)         & -- & \checkmark         & 64.26 & 0.0107 & 294.94 \\
HQQ (4b)         & -- & \checkmark         & 64.64 & 0.0071 & 248.76 \\
Quanto (4b)         & -- & \checkmark         & 65.42 & 0.0069 & 294.40 \\
LQA-Lite(Ours)         & -- & \checkmark         & 65.69 & 0.0071 & 265.80 \\

BnB (8b)         & -- & \checkmark         & 65.46 & 0.0096 & 278.49 \\
HQQ (8b)         & -- & \checkmark         & 65.52 & 0.0070 & 266.75 \\
Quanto (8b)         & -- & \checkmark         & 65.76 & 0.0068 & 296.54 \\
LQA(Ours)         & -- & \checkmark         & 65.76 & 0.0068 & 296.38 \\
\midrule
BnB (4b)         & \checkmark & \checkmark         & 59.99 & 0.0223 & 189.24 \\
HQQ (4b)         & \checkmark & \checkmark         & 62.91 & 0.0086 & 132.67 \\
Quanto (4b)         & \checkmark & \checkmark         & 60.42 & 0.0372 & 290.13 \\
LQA-Lite(Ours)         & \checkmark & \checkmark         & 63.98 & 0.0081 & 148.31 \\

BnB (8b)         & \checkmark & \checkmark         & 65.50 & 0.0573 & 222.34 \\
HQQ (8b)         & \checkmark & \checkmark         & 65.38 & 0.0116 & 202.24 \\
Quanto (8b)         & \checkmark & \checkmark         & 65.69 & 0.0212 & 291.52 \\
LQA(Ours)         & \checkmark & \checkmark         & 65.56 & 0.0114 & 195.76 \\
\bottomrule
\end{tabular}
}
\vspace{-1.5em}
\label{tab:ablation-main}
\end{table}

\noindent
Under the 8-bit and 4-bit quantization (Table~\ref{tab:ablation-main}), we observe that our proposed LQA and LQA-Lite configurations consistently outperform baseline quantization methods in terms of accuracy–efficiency trade-offs. For the vision encoder only quantization setting, LQA-Lite achieves the highest accuracy at both 4-bit (64.13\%) and 8-bit (65.89\%) while maintaining the lowest latency (0.0089s and 0.0113s) and memory footprint (184.64MB and 231.08MB). Compared to baseline methods, LQA and HQQ consistently offer lower latency and memory usage than BnB and Quanto, while matching or exceeding their accuracy.
In the text encoder only quantization setting, LQA and LQA-Lite again lead in accuracy, reaching 65.76\% and 65.69\% at 8-bit and 4-bit respectively, while maintaining lowest latency and low-to-moderate memory usage. Quanto performs competitively in accuracy and runtime latency but incurs noticeably higher memory costs, especially in the 4-bit setting. BnB and HQQ perform similarly in terms of latency under 8-bit, while HQQ is more memory-efficient. Under 4-bit setting, HQQ out forms both BnB and Quanto in terms of latency and memory usage.
In the setting of both encoders quantized, the gap between quantization methods becomes more evident. LQA-Lite achieves the best performance at 4-bit with 63.98\% accuracy and just 0.0081s latency, while also consuming significantly less memory (148.31 MB) compared to Quanto (290.13 MB) and BnB (189.24 MB). At 8-bit setting, LQA achieves the highest accuracy (65.56\%) and remains the most memory-efficient among all methods. These results validate the effectiveness of our hybrid quantization strategy, which selectively applies structured precision control to each modality.

\begin{table}[t]
\centering
\caption{Ablation study of various low-bit quantization methods evaluated on different encoder configurations.}
\resizebox{\columnwidth}{!}{
\begin{tabular}{lccccc}
\toprule
\textbf{Method} & \textbf{Vision} & \textbf{Text} & \textbf{Accuracy (\%)} & \textbf{Latency (sec)} & \textbf{Memory (MB)} \\
\midrule
HQQ (1b) & \checkmark & -- & 2.87 & 0.0273 & 199.71 \\

HQQ (2b) & \checkmark & -- & 46.89 & 0.0229 & 211.83 \\
Quanto (2b) & \checkmark & -- & 10.09 & 0.0319 & 237.72 \\

HQQ (3b) & \checkmark & -- & 61.85 & 0.0343 & 215.25 \\
\midrule
HQQ (1b) & -- & \checkmark & 3.13 & 0.0089 & 257.86 \\

HQQ (2b) & -- & \checkmark & 61.28 & 0.0077 & 262.60 \\
Quanto (2b) & -- & \checkmark & 32.68 & 0.0060 & 239.53 \\

HQQ (3b) & -- & \checkmark & 63.91 & 0.0093 & 262.60 \\
\midrule
HQQ (1b) & \checkmark & \checkmark & 3.02 & 0.0285 & 149.46 \\

HQQ (2b) & \checkmark & \checkmark & 42.54 & 0.0195 & 164.35 \\
Quanto (2b) & \checkmark & \checkmark & 9.50 & 0.0144 & 102.22 \\

HQQ (3b) & \checkmark & \checkmark & 60.59 & 0.0321 & 170.66 \\
\bottomrule
\end{tabular}
}
\label{tab:ablation-lowbit}
\vspace{-1.5em}
\end{table}

\noindent
To further assess the limits of quantization, we explore low-bits precision under 3-bit, 2-bit, and 1-bit quantization (Table~\ref{tab:ablation-lowbit}). As expected, extreme quantization leads to substantial performance degradation. For the vision encoder, HQQ-3bit retains 61.85\% accuracy, while HQQ-2bit drops to 46.89\%, and HQQ-1bit collapses to just 2.87\%. Quanto-2bit also performs poorly with only 10.09\% accuracy. In the text encoder, HQQ-3bit again performs best (63.91\%), while Quanto-2bit and HQQ-2bit exhibit moderate degradation (32.68\% and 61.28\%, respectively). However, at 1-bit, HQQ fail to preserve semantic understanding, dropping below 4\% accuracy.

\noindent
In the both encoders quantized setting, results follow the same trend. HQQ-3bit remains usable (60.59\%), but 2-bit and 1-bit quantization result in steep performance loss, with Quanto-2bit achieving only 9.50\% and HQQ-1bit falling to 3.02\%. Despite slightly lower memory usage in these low-bit settings, the performance degradation outweighs the gained efficiency. These results highlight the limitations of aggressive quantization and suggest that 3-bit quantization may represent the lowest feasible precision for practical use without severe accuracy loss. However, when compared to HQQ-4bit, HQQ-3bit is still outperformed in terms of accuracy, latency, and memory footprint, indicating that 4-bit remains the more balanced and effective configuration.

\noindent
Overall, the results demonstrate that our LQA and LQA-Lite configurations offer superior trade-offs between accuracy, latency, and memory consumption. They effectively outperform standard quantization methods under both moderate (4/8-bit) and aggressive (less than 3-bit) precision settings across all encoder configurations.

%% file: sections/Appendix.tex
\clearpage
\appendix
\section*{Appendix}
\addcontentsline{toc}{section}{Appendix} 
\section{Datasets}
\begin{itemize}
\item \textbf{CIFAR10}~\cite{cifar}: A well-known dataset used in vision language benchmark~\cite{hendrycks2019benchmarking, maharana2024batclip, osowiechi2024watt, kumar2024bimodaltta}, that consist of 60,000 $32\times32$ color images in 10 classes, with evenly 6,000 images per class. Each class it-self are completely mutually exclusive to each other. By default, the 6,000 of each class is randomly split into 5,000 training data and 1,000 test data.
\item \textbf{CIFAR100}~\cite{cifar}: An extension to CIFAR10, sharing the same image resolution and format, but has 100 mutually exclusive classes, each with 600 images. It follows the same 5:1 training-test ratio as CIFAR10, resulting in 500 training data and 100 test data per class. The increased number of classes from 10 to 100 making it significantly more challenging to models due to finer label granularity and higher inter-class similarity. Along with CIFAR10, CIFAR100 serves as a more fine-grained benchmark that stresses a model’s scalability and robustness under increased classification complexity.
\item \textbf{Caltech101}~\cite{caltech101}: A standard image classification dataset consisting of 9,146 images across 101 object categories and one background clutter class. Each image is labeled with a single object and about 40 and 800 images per category, with most classes having around 50 images. All images within the uniform resolution, but vary in shape and background, offering moderate variability. Caltech101 is widely used to evaluate models’ ability to generalize under limited training data and high inter-class diversity~\cite{fei2007learning, koch2015siamese, vinyals2016matching}.
\item \textbf{Oxford–IIIT Pet}~\cite{oxford_pets}: A fine-grained dataset of 7,349 images of 37 category of cat and dog breeds. Each image is annotated with a class label, tight bounding box (ROI) around head and pixel level tri map segmentation. They are largely variate in scale, pose and lighting, making the dataset useful for both classification and segmentation tasks. Class sizes are roughly balanced at 200 images per class.
\item \textbf{Describable Textures Dataset (DTD)}\cite{dtd}: A evolving database of 5,640 texture images collected from Google and Flickr, annotated with 47 human-centric attributes. Each category has 120 images, split into 40:40:40 for training, validation, and testing. DTD is widely used to assess a model’s ability to recognize fine-grained, non-object visual concepts.
\item \textbf{UFC101}~\cite{ucf101}: A large-scale action recognition benchmark dataset containing 13,320 video clips collected from YouTube. It has 101 realistic action categories grouped into 25 groups, with each consist of 4-7 videos of an action. Each Clip vary in camera motion, object appearance, pose, object scale, viewpoint, lighting, and background, serves as a challenging test bed for spatio-temporal generalisation. The training and test dataset are split into the ratio of 3:1.
\item \textbf{ImageNet Adversarial (ImageNet A)}\cite{hendrycks2021imgA}: A curated set of 7,500 unmodified natural photographs drawn from 200 ImageNet-1k classes but specifically chosen to fool existing ImageNet classifiers. It is created with a simple adversarial filtration technique to offer limited spurious cues, therefore serves as an most challenging adversarial benchmark that tests a model’s ability to handle rare viewpoints, occlusions, and atypical object-background contexts.
\end{itemize}


\section{LQA Configurations}
In this section, we detail the implementation regarding the proposed Selective Hybrid Quantization (SHQ) framework and Q-TTA adaptation strategy. Firstly, we specify the layer-wise precision settings for both the vision and text encoders, and then outline cache configurations utilized during test-time adaptation.

\subsection{LQA}
All Transformer weight layers of the vision encoder (\(E_V\)), such as self-attention and MLP modules are quantized using a Hessian-aware quantizer (HQQ) with 8-bit precision and a group size of 128. To ensure the preservation of numerical stability, the final \textit{LayerNorm} and \textit{Projection} layer are selected to remain in float16 to minimize dequantization overhead and distortion. For text encoder (\(E_T\)), all weight layers are quantized with a symmetric uniform quantizer (Quanto) at 8-bit precision, while the \textit{Projection} layer is kept in float16 precision.

\subsection{LQA-Lite}
In LQA-Lite, the weight layers in vision encoder are quantized with HQQ at 4-bit precision in a group size 64, selectively retaining the final \textit{LayerNorm} layer in float16. The text encoder employed HQQ and quantized all layers in 8-bit precision and a group size of 256, with the \textit{Projection} layer left in float16.

\subsection{Q-TTA Configuration}
\begin{table}[h]
\centering
\caption{Positive Cache Configuration for each dataset.}
\resizebox{\columnwidth}{!}{
\begin{tabular}{lcccc}
\toprule
\textbf{Dataset} & \textbf{Enabled} & \textbf{Shot Capacity} & \textbf{Alpha} & \textbf{Beta} \\
\midrule
CIFAR-10      & True & 3 & 1.0 & 8.0 \\
CIFAR-100     & True & 3 & 1.0 & 8.0 \\
Caltech-101   & True & 3 & 5.0 & 5.0 \\
Oxford        & True & 3 & 2.0 & 7.0 \\
DTD           & True & 3 & 2.0 & 3.0 \\
UCF-101       & True & 3 & 3.0 & 8.0 \\
ImageNet-A    & True & 3 & 2.0 & 5.0 \\
\bottomrule
\end{tabular}
}
\label{tab:poscache}
\end{table}

\begin{table}[h]
\centering
\caption{Negative-cache hyper-parameters shared for all datasets.}
\resizebox{0.8\columnwidth}{!}{
\begin{tabular}{lccc}
\toprule
\textbf{Parameter} & \textbf{Value} & \textbf{Lower} & \textbf{Upper} \\
\midrule
Enabled            & True  & --   & --   \\
Shot Capacity      & 2     & --   & --   \\
Alpha              & 0.117 & --   & --   \\
Beta               & 1.0   & --   & --   \\ \midrule
Entropy Threshold  & --    & 0.20 & 0.50 \\
Mask Threshold     & --    & 0.03 & 1.00 \\ 
\bottomrule
\end{tabular}
}
\label{tab:negcache}
\end{table}

In the setting of Q-TTA, the mechanism employs both positive and negative caches to adapt the model during test-time without gradient computations. Due to that, it is essential to carefully tune cache-related hyperparameter, including \textbf{shot capacity} (the number of samples kept in cache), \textbf{$\alpha$} (influence of positive cache), \textbf{$\beta$} (influence of negative cache) and filtering thresholds (e.g., entropy and mask thresholds), to ensure effective pseudo label reliability. As these parameters directly influence the quality of the adaptation process, careful finetuning is required to practice. Take into this consideration, we utilized the existing configuration for each dataset from official TDA implementation~\cite{Karmanov2024TDARepo}, as our Q-TTA build upon TDA and they both share the similar core architecture. These configuration has been proved effective to be effective in prior experiments~\cite{karmanov2024TDA}. 

\noindent
As demonstrated in Table~\ref{tab:poscache} and Table~\ref{tab:negcache}, \textbf{positive cache} adopt dataset-specific strategy to accommodate the confident samples across different domains. For instance, classes from Caltech-101 are more likely benefit from a higher $\alpha$ than CIFAR-10 and CIFAR-100, as have more abstract categories and less confident predictions. A higher $\alpha$ introduces more influence from positive cache to help correctness. Similarly for $\beta$. These values are dataset variant, therefore are configured separately. In contrast, negative cache are meant to be more conservative and noise-tolerant to avoid harming predictions. To main stability, uniform configuration are applied to all datasets. For instance, less confident samples (limited to 2 per class) are stored in negative cache. Furthermore, only extremely low probability (less than 0.03) labeled as negative, which reduces the risk of mistakenly suppressing plausible classes. In addition, only exemplars with moderate entropy are allowed in negative cache, as samples with entropy below 0.2 may be Overconfident while those above 0.5 are often uncertain and noisy-prone. These conservative values ensure that \textbf{negative cache} remains generalizable and robust across diverse domains.


\section{Hyperparameters of Gradient-based Methods}
Tent and EATA are optimised only on normalisation parameters using plain SGD with learning rate $1.6{\times}10^{-5}$, while SAR uses SAM-SGD with doubled rate $3.1{\times}10^{-5}$. The entropy cut-off for adaptation is shared by SAR and EATA as $2.76$; EATA also removes redundant inputs (\texttt{d\_margin} $=0.05$) and guards forgetting with a Fisher penalty (\texttt{fisher\_alpha} $=2000$). SAR uses an additional recovery trigger to reset the model when the EMA of this loss drops below \texttt{reset\_constant\_em} $=0.2$. TDA skips gradient updates, relying on cache dynamics: the positive cache stores \texttt{shot\_capacity} $=3$ instances re-weighted by \texttt{alpha} $\in\{1.0,2.0,5.0\}$ and \texttt{beta} $\in\{3.0,5.0,8.0\}$; the negative cache stores 2 shots with \texttt{alpha} $=0.117$, \texttt{beta} $=1.0$, and only activates for entropy in $[0.2,\,0.5]$, per-class probabilities in $[0.03,\,1.0]$. These settings follow the original paper and adhere to standard configurations for the target datasets.

\section{Quantization Trade-offs Analysis}
To evaluate the efficiency of different quantization methods beyond accuracy, we analyze their runtime latency and memory consumption across varying group sizes and bit-width settings. 

\begin{figure} [h]
    \centering
    \includegraphics[width=1\linewidth]{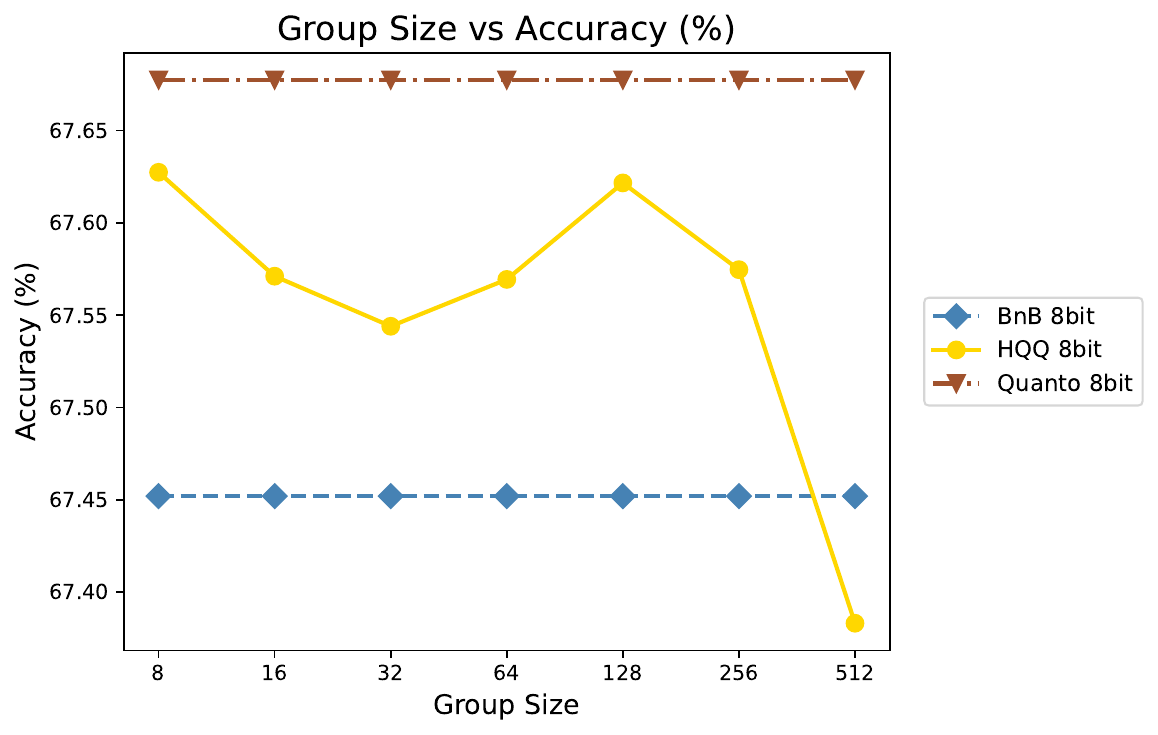}
    \caption{Average Accuracy of Different Quantization Method in 8-bit.}
    \label{fig:Average Accuracy of Different Quantization Method in 8-bit}
\end{figure}

\begin{figure}[h]
    \centering
    \includegraphics[width=1\linewidth]{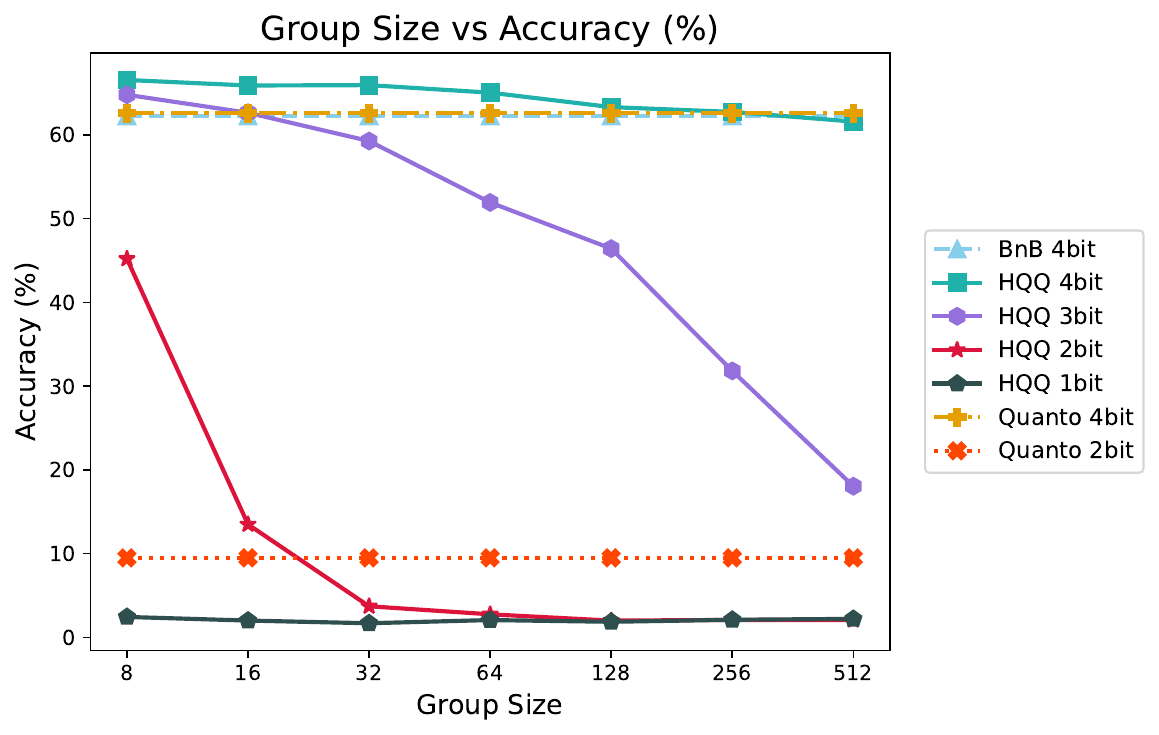}
    \caption{Average Accuracy of Different Quantization Method in other-bits.}
    \label{fig:Average Accuracy of Different Quantization Method in other-bits}
\end{figure}

\noindent
As demonstrated in Figure~\ref{fig:Average Accuracy of Different Quantization Method in 8-bit}, all methods maintain relatively stable and similar performance in 8-bit quantization, with 67.69, 67.63 and 67.38 respectively. As group size configuration require flexible control over the granularity of quantization, only HQQ implemented with different group size, other method like Quanto and BnB stayed the same uniform quantization strategy, resulting in consistent behavior across group size configurations. Quanto-8bit achieved the highest accuracy, followed by HQQ and BnB. In group size wise comparison, HQQ-8bit exhibits highest accuracy at group size 8 and 128, slightly dropped at 256 and 512. In lower-bits comparison (Figure~\ref{fig:Average Accuracy of Different Quantization Method in other-bits}), three 4-bit method showed the comparable result, with HQQ achieved the highest accuracy. However, accuracy generally decreases as the bit-width decreases. HQQ 4bit maintains competitive accuracy until group size 64, but HQQ 2bit and especially HQQ 1bit suffer significant degradation, particularly with larger group sizes. Similar degradation observed on other extreme low-bit method, like Quanto 2 bit. This indicates model are able to maintain their performance at 8bit most of the time, and does not impact much by increasing group size; 4 bit quantization are able to keep a majority of performance, but less robust after group size increased (64); lower low-bits (3b), especially 2b and 1b significantly loss model performance and most sensitive with increasing group size; for extreme low bit, like 2b, 1b, the model likely to be corrupted after quantization and fail to retain their predictive capabilities.

\begin{figure}[h]
    \centering
    \includegraphics[width=1\linewidth]{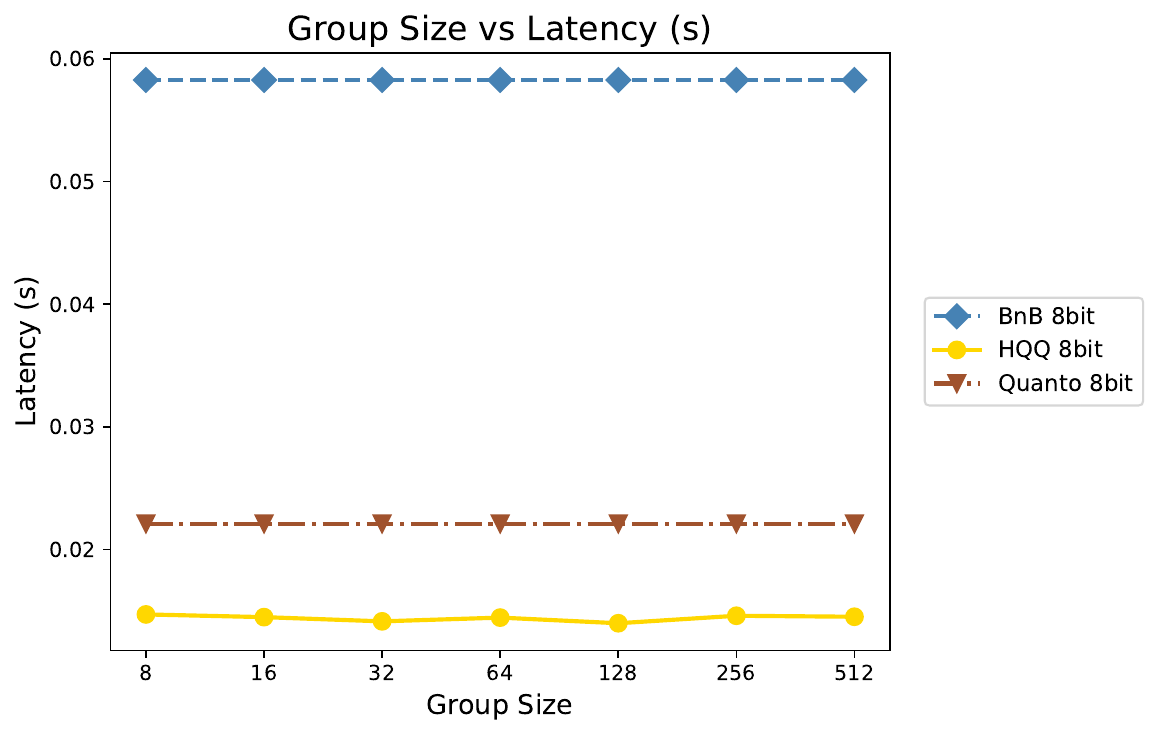}
    \caption{Average Latency of Different Quantization Method in 8-bit.}
    \label{fig:Average Latency of Different Quantization Method in 8-bit}
\end{figure}

\begin{figure}[h]
    \centering
    \includegraphics[width=1\linewidth]{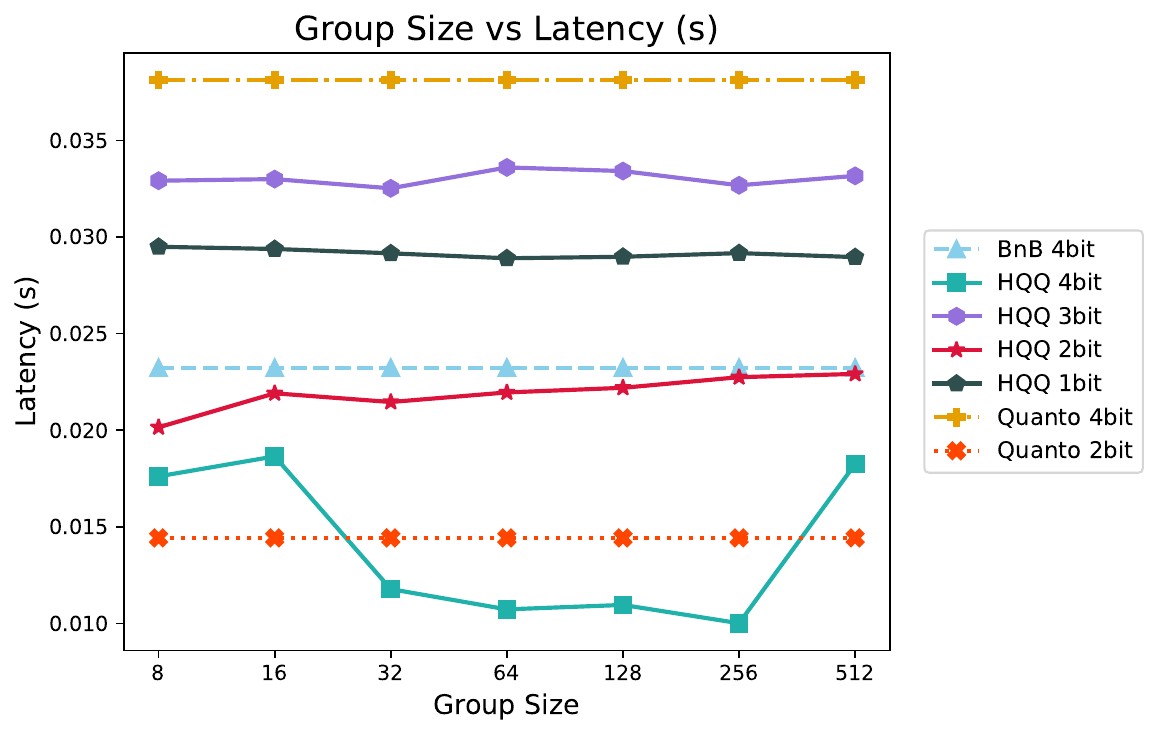}
    \caption{Average Latency of Different Quantization Method in other-bits.}
    \label{fig:Average Latency of Different Quantization Method in other-bits}
\end{figure}

\noindent
In Latency analysis, as shown on Figure~\ref{fig:Average Latency of Different Quantization Method in 8-bit}, among 8-bit methods, BnB-8bit incurs the highest latency, while HQQ-8bit achieves the lowest and remains consistently faster than Quanto-8bit across all group sizes. In group size comparison, 128 showed the lowest latency in HQQ 8 bit. In low-bits settings (Figure~\ref{fig:Average Latency of Different Quantization Method in other-bits}), Quanto-2bit offer the lowest latency, reflecting their efficient computation in low-bit, while Quanto-4bit incurs highest latency. HQQ-4bit is also efficient, with slightly higher latency at low group size 8 and 16, but surpass Quanto 2bit at 32 and achieved lowest value at 256. However, for extreme low-bit quantization, HQQ 3bit, HQQ 2bit and HQQ 1bit experience higher latency, likely due to the computational overhead of extremely fine-grained quantization. These results suggest that HQQ-4bit with a moderate group size offers the best trade-off between latency and quantization granularity, while lower-bit configurations introduce diminishing returns due to processing overhead.

\begin{figure}[h]
    \centering
    \includegraphics[width=1\linewidth]{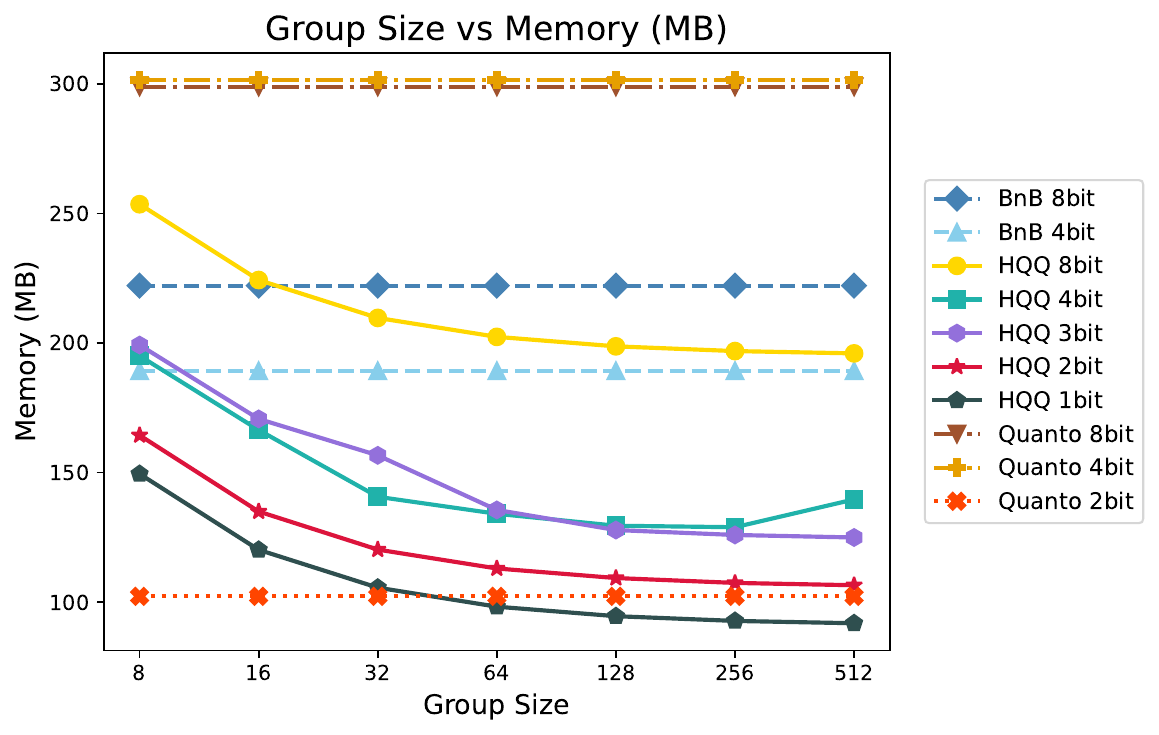}
    \caption{Average Latency of Different Quantization Method.}
    \label{fig:Average Latency of Different Quantization Method}
\end{figure}

\noindent
Memory consumption reveals clear trends as presented in Figure 9. HQQ-based methods show gradually decreasing memory usage with increasing group size, owing to improved compression from larger quantization groups (lower group size, more group there are and more values need to store). In contrast, Quanto maintains a relatively high and flat memory footprint across all settings, especially in 8-bit configurations, due to the lack of group-wise compression. BnB also maintains a nearly constant memory profile, as it internally uses fixed block-wise quantization.